\documentclass{article}

\usepackage[final]{corl_2020} % Uncomment for the camera-ready ``final'' version.

\usepackage{amssymb}
\usepackage{pifont}
\usepackage{amsmath}
\usepackage{caption, subcaption}
\usepackage{graphicx}
\usepackage{appendix}
\usepackage{etoolbox}
\usepackage{hyperref}
\usepackage{etoolbox}
\usepackage{wrapfig}
\newcommand{\cmark}{\ding{51}}%
\usepackage{xcolor}
\usepackage{booktabs}
\usepackage{arydshln}
\hypersetup{
    colorlinks=true,
    linkcolor=blue,
    filecolor=magenta,      
    urlcolor=cyan,
    anchorcolor=blue,
}
\makeatletter
\newcommand*\bigcdot{\mathpalette\bigcdot@{.5}}
\newcommand*\bigcdot@[2]{\mathbin{\vcenter{\hbox{\scalebox{#2}{$\m@th#1\bullet$}}}}}
\makeatother

\makeatletter
\def\thickhline{%
  \noalign{\ifnum0=`}\fi\hrule \@height \thickarrayrulewidth \futurelet
   \reserved@a\@xthickhline}
\def\@xthickhline{\ifx\reserved@a\thickhline
               \vskip\doublerulesep
               \vskip-\thickarrayrulewidth
             \fi
      \ifnum0=`{\fi}}
\makeatother
\newlength{\thickarrayrulewidth}
\title{Interaction-Based Trajectory Prediction \\ Over a Hybrid Traffic Graph}

\author{
  Sumit Kumar, Yiming Gu, Jerrick Hoang, Galen Clark Haynes, Micol Marchetti-Bowick\\[3pt]
  Uber Advanced Technologies Group, Pittsburgh, PA\\[3pt]
  \texttt{\{sumitk,yiming,jhoang,gch,mmarchettibowick\}@uber.com}\\
}

\begin{document}

\maketitle

%===============================================================================

\begin{abstract}
Behavior prediction of traffic actors is an essential component of any real-world self-driving system. Actors' long-term behaviors tend to be governed by their {interactions} with other actors or traffic elements (traffic lights, stop signs) in the scene. To capture this highly complex structure of interactions, we propose to use a hybrid graph whose nodes represent both the traffic actors as well as the static and dynamic traffic elements present in the scene. The different modes of temporal interaction (e.g., \emph{stopping} and \emph{going}) among actors and traffic elements are explicitly modeled by graph edges. This explicit reasoning about discrete interaction types not only helps in predicting future motion, but also enhances the {interpretability} of the model, which is important for safety-critical applications such as autonomous driving. We predict actors' trajectories and interaction types using a graph neural network, which is trained in a semi-supervised manner. 
We show that our proposed model, TrafficGraphNet, achieves state-of-the-art trajectory prediction accuracy while maintaining a high level of interpretability.
\end{abstract}


%==============

\section{Introduction}

% \begin{wrapfigure}{r}{0.5\textwidth}
%  \begin{center}
%     \includegraphics[width=0.48\textwidth]{figures/introduction_picture.png}
%  \end{center}
%  \caption{Caption}
% \end{wrapfigure}

Autonomous driving is one of the most exciting real-world applications of artificial intelligence because it has the potential for enormous societal benefit. In order to plan a safe and comfortable trajectory to its destination, a self-driving vehicle (SDV) must reason about the future motion of all other actors in the scene. This reasoning must occur over a sufficiently long time horizon to allow the SDV to plan out complex maneuvers, such as making a lane change in dense traffic or navigating a 4-way stop intersection with other vehicles. Future motion prediction is particularly challenging in busy scenes with many interacting agents, including the SDV itself.

Over short-term horizons (e.g., 1 to 3 seconds), a vehicle's future motion is heavily constrained by vehicle dynamics, which means it is primarily driven by its current motion. In contrast, over longer-term horizons (e.g., 6 to 10 seconds), vehicle behavior is primarily driven by its intended destination and its interactions with other actors, traffic signals, and the environment. In particular, we observe that the long-term future trajectories of traffic actors usually follow a set of underlying discrete interaction decisions (such as ``A yields to B'' vs.~``B yields to A''). Furthermore, we recognize that these N-way negotiations happen not only among traffic actors but also between traffic actors and traffic elements (such as ``A stops for a stop sign'' vs.~``B goes when the light turns green''). Identifying and categorizing these behaviors is not only useful for predicting future motion, but also for {describing} or {explaining} a traffic scene in a human-understandable way. 

% TODO: we mention `two-stage` hybrid graph, but never mention what `two-stage` mean?
Starting from these key observations, we propose a novel method for long-term, interaction-based vehicle trajectory prediction. We build a two-stage hybrid graph network, which we describe as ``hybrid'' because its nodes comprise both traffic actors (including all non-parked vehicles in the scene) and traffic elements (including traffic lights, stop signs, and yield signs). The edges of this graph capture semantically meaningful interaction types, allowing information to flow through the graph in an interpretable manner. 

With this architecture, we are able to demonstrate better performance over existing state-of-the-art methods both on our internal real-world driving dataset as well as on the open-source nuScenes dataset. 
Furthermore, we demonstrate the interpretability of the predictions made by our model.  By manually perturbing graph edges between pairwise actors (e.g. yield $\rightarrow$ go) or by changing traffic light states in the graph (e.g. red $\rightarrow$ green), we produce new trajectory predictions corresponding to what actors \emph{would have done} in these hypothetical situations.  As the interaction types produced by our model correspond to semantically distinct modes of behavior, we are able to use these as interpretable explanations of the resulting trajectory predictions, a key ingredient for developing and maintaining a safety-critical AI system such as a self-driving car.

\section{Related Work}

Behavior prediction is an essential part of the autonomous driving task. %Consequently, there has been significant research dedicated to predicting future behaviour and motion of road users. 
Since some of the earliest seminal work on behavior forecasting~\cite{helbing1995social, kaempchen2004imm, barth2008will, kitani2012activity}, there has been significant research dedicated to predicting future behaviour and motion of road users. 

%One line of research is to frame the behavior prediction as a Partially-Observable Markov Decision Process (POMDP)~\cite{kitani2012activity}. Behavior prediction is done by Inverse Reinforcement Learning to recover the actor's policies. %Furthermore, behavior prediction has been included as part of a larger end-to-end learning system that performs multi-task learning on multiple autonomous driving objectives. One example is Fast-And-Furious (FAF)~\cite{luo2018fast}, where raw sensor data was used as an input to jointly reason about detection, tracking, and motion forecasting. Based on FAF, IntentNet~\cite{casas2018intentnet} attempts to reason about longer term trajectories by predicting additional high-level intents for traffic actors. MultiXNet~\cite{djuric2020multinet} further pushes the boundary by proposing a two-stage system, where in the second stage multi-modal uncertainty-aware trajectories are proposed and refined.

\textbf{Context Representation}: One important aspect of behavior prediction is to learn useful representations of the contextual information necessary for predicting actors' future behaviors. Rasters have been used extensively to encode the information about scene context such as maps and traffic signals. Works such as RasterNet~\cite{chou2019predicting, djuric2018motion, djuric2018short}, CoverNet~\cite{phan2020covernet}, and ChauffeurNet~\cite{bansal2018chauffeurnet} encode the map and surrounding information in a multi-channels bird's eye view raster and use CNNs to extract information. VectorNet~\cite{gao2020vectornet} takes a step further by encoding the context in graphs. In particular, elements in a traffic scene such as lane boundaries and agents' trajectories are represented as vectors, which in turn are represented as polyline graphs and interaction graphs. To further enhance the context representation, behavior prediction has been included as part of a larger end-to-end learning system that performs multi-task learning on both perception and prediction. One example is Fast-And-Furious (FAF)~\cite{luo2018fast}, where raw sensor data was used as an input to jointly reason about detection, tracking, and motion forecasting. Based on FAF, IntentNet~\cite{casas2018intentnet} attempts to reason about longer-term trajectories by predicting additional high-level intents for traffic actors. MultiXNet~\cite{djuric2020multinet} further pushes the boundary by proposing a two-stage system, where in the second stage, multi-modal uncertainty-aware trajectories are proposed and refined.

\textbf{Generative Models}: While context representation is important to capture information from the surrounding scene, estimating the full distribution over actors' future behaviors is equally essential due to the inherent ambiguity of the trajectory prediction task. One line of research in this direction is to use generative models, including %can offer accurate descriptions of the resulting trajectory distributions, which is equally essential. One line of research in this direction is to use 
Generative Adversarial Networks such as SocialGAN~\cite{gupta2018social}, Multi-Agent Tensor Fusion (MATF)~\cite{zhao2019multi}, and SoPhie~\cite{sadeghian2019sophie}; Variational Autoencoders, such as DESIRE~\cite{lee2017desire} and R2P2~\cite{rhinehart2018r2p2}; and Normalizing Flows, such as PRECOG~\cite{rhinehart2019precog}. These models ensure the output distribution of the model approaches the true distribution. 

\textbf{Interaction Modeling}: There has been an increasing amount of work on modeling the interaction aspect of behavior prediction in a multi-agent setting. In earlier works such as SocialLSTM~\cite{alahi2016social} and Convolutional Social Pooling (CSP)~\cite{deo2018convolutional}, interaction among smart agents are implicitly modeled by the ``social pooling'' operation.
SoPhie~\cite{sadeghian2019sophie} extends on social pooling by introducing various attention mechanisms to model the agent-agent interaction and agent-environment interaction through the ``social attention'' and ``physical attention'' modules. Different from the social pooling among RNN hidden states, Multi-Agent Tensor Fusion (MATF)~\cite{zhao2019multi} jointly encodes both the context and agents' past trajectories in a single ``multi-agent tensor'' and recurrently decodes the multi-agent tensor through convolutional fusion. Additionally, SpAGNN~\cite{casas2019spatially} models the multi-agent system using a Graph Neural Network, where the node features are derived from the combination of cropped map information and the upstream features extracted from the sensor data. 

In the above works, although interaction is considered implicitly through aggregation operations such as social pooling, social attention, convolutional fusion, and graph network message passing, the interactions are not explicitly supervised. 
For example, we can infer the feature importance by examining the social attention layers, but it remains hard to interpret how these features are used for downstream prediction. IntentNet~\cite{casas2018intentnet} moves one step further in terms of interpretability by performing ``intent prediction'' on actors. 
However, the intents are limited to individual isolated actions, rather than interactions between actors. In this paper, we push interpretability further by introducing explicit interaction prediction between pairs of actors. 
We introduce the notion of an hybrid graph whose nodes represent both traffic actors and traffic elements, and edges represent the explicit pairwise interactions among them. This allows us to accurately model traffic actors' behavior in difficult scenarios involving complex multi-way interactions. 
Our proposed model, TrafficGraphNet, by explicitly reasoning about the discrete interaction between both actors and elements, is able to achieve state-of-the-art prediction performance while maintaining a high level of interpretability.

\section{Method}
\begin{figure}
    \centering
    \includegraphics[width=\linewidth]{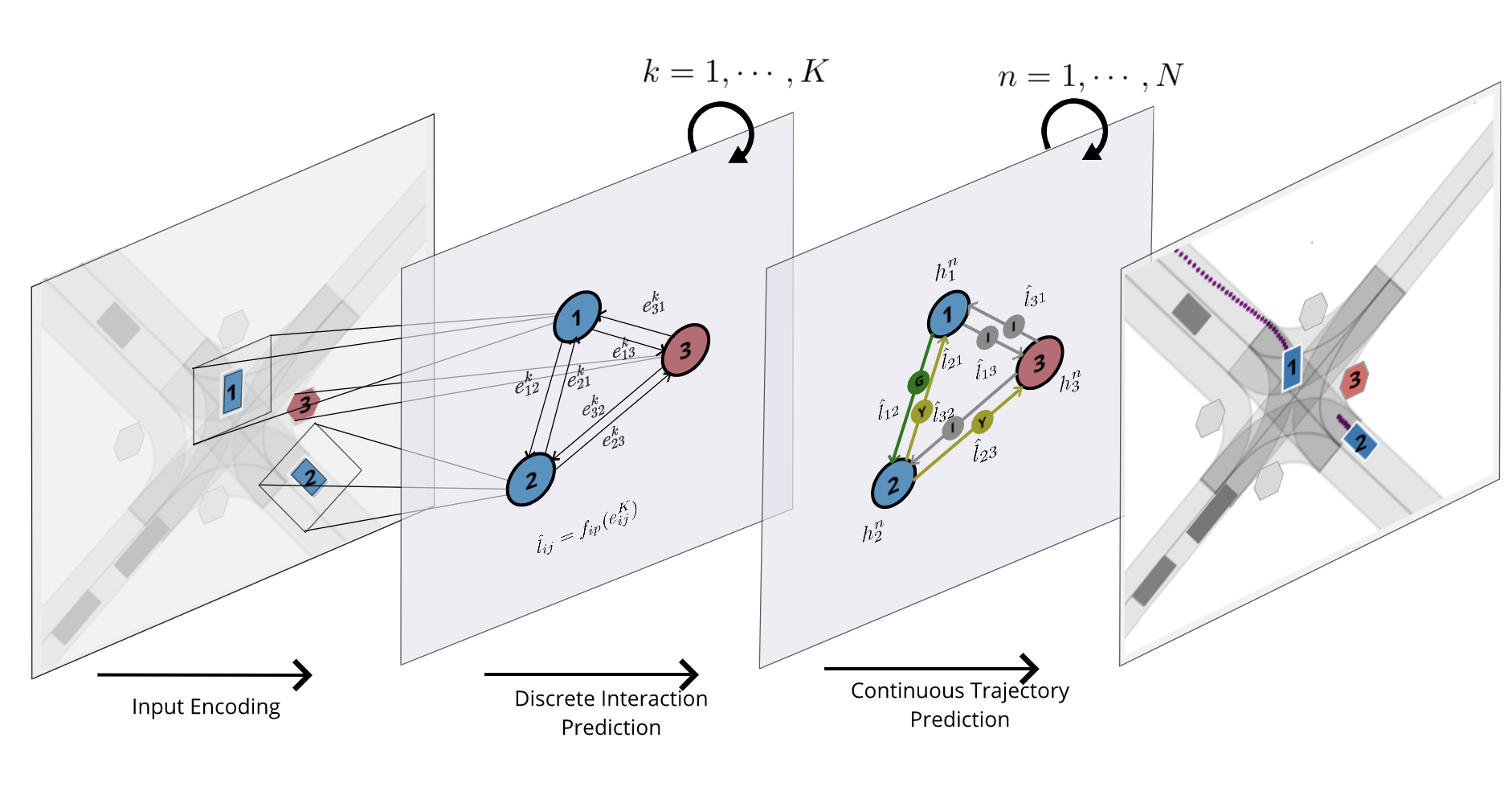}
    \caption{An overview of the TrafficGraphNet (TGN): In the first stage (Input Encoding), we extract information (such as location, nearby map rasters, and history) about actors and traffic elements into graph nodes, and connect nodes with edges to form our graph structure. In the second stage (Discrete Interaction Prediction), we use a GNN to predict actor-actor and actor-signage interaction on the graph edges. In the third stage (Continuous Trajectory Prediction), given the graph and the edge predictions, we predict continuous trajectories for all actors.} % Text
    \label{fig:method}
\end{figure}

In this section, we provide a description of the different components of our model, which are illustrated in Figure \ref{fig:method}. We first describe the hybrid graph formulation and input encoding in Section \ref{sec:hybrid_graph}. We then describe the discrete interaction prediction module and traffic light state prediction module in Sections \ref{sec:interaction_prediction} and \ref{sec:tl_prediction}. Finally, we describe continuous trajectory prediction in Section \ref{sec:traj_prediction}. %An overview of our model is shown in Figure \ref{fig:method}.

\subsection{Hybrid Graph Representation} \label{sec:hybrid_graph}
\vspace{-.2cm}
%TODO: we did not talk about tl state prediction at all before its section?
% graph nodes and edges, graph building
We represent the traffic scene as a hybrid graph with traffic actors and traffic elements as nodes. We consider all non-parked vehicles as traffic actors. We also consider $6$ types of traffic elements: stop signs, yield signs, red traffic lights, green traffic lights, yellow traffic lights, and unknown traffic lights. Unknown traffic lights are those whose current state cannot be observed due to reasons like occlusion, glare, poor visibility, etc. A directional edge $(i,j)$ in the graph serves as a medium for propagating information from source node $i$ to destination node $j$. All the actor nodes are connected to each other, as done in other prior works~\cite{casas2019spatially, lee2019joint}. An edge exists between a traffic element and an actor if the distance between them is less than a predefined threshold of $25m$. This threshold is set based on the observation that actors usually interact with only those traffic elements that are in their close proximity. This graph representation enables us to utilize the inherent structural information present in the scene, reason about the discrete interaction type for each edge, and predict future trajectories conditioned on the interaction type for any number of actors in the scene.

% TODO: put one line justify this design choice, have something like this:
% This graph building heuristic ensures a generic scene representation where any two vehicle nodes can share information among each other while also keeping the computational cost in check.

% node input: type, history, raster
The nodes in our hybrid graph can be of $7$ different types: $1$ vehicle type actor node and $6$ traffic element nodes as described above. The node type information for node $i$ is represented in one-hot form $v_i$. We assume each node has access to its own past motion states, like position and velocity for a duration of $2s$ at $2Hz$ frequency. Since traffic elements are stationary, their past states consist of fixed position and zero velocity. Furthermore, each node is also provided with a $40m \times 40m$ map raster at a resolution of $0.2m$/pixel. Each node has a field of view of $30m$ in the front, $10m$ in the back, and $20m$ on each side. Different elements of the map, like lane boundaries, left turn region, right turn region, and motion paths are represented in different channels of the raster. The raster provides necessary scene information, like driveable areas, intersection, etc. to each node.

%TODO: put in appendix:
% $f_s$ is parameterized as an RNN with a GRU cell. 
% We parameterize $f_m$ by a $5$-layer CNN followed by a single FC layer.
% We set number of GN layers $K$ to be $2$. 
% Here also, we used $K=2$ layers. 
A state encoder module $f_s$ encodes the past trajectory $s_i^{t-H:t}$ for each actor as $h^s_i = f_s\left(s_i^{t-H:t}\right)$ where $H$ is the number of history frames. The past trajectory is transformed into the node-frame, whose origin is at the position of the node and the y-axis aligns with the heading direction. We compute the heading for traffic elements by calculating the vector from the start location to the end location of their control region. %, which we can access in our HD maps. 
If a traffic element is part of multiple control regions, we take the mean vector as the heading direction. For example, a stop sign at a $4$-way stop intersection has $3$ control regions, one each for going straight, left turn, and right turn. The start point of the control region is the location of the traffic element itself and the end point is the location after crossing the intersection, which is different for the $3$ cases. The scene map raster $m_i$ for a node $i$ is also transformed to its own frame. %Transforming both, past trajectory and raster to node frame, simplifies the input representation for the model and accelerates model’s convergence. 
A map encoder module $f_m$ encodes the map raster for each actor as $h^m_i = f_m\left(m_i\right)$. A fuse layer $f_l$ then combines the node type, state, and map encodings together as 
% Old ^
% If a traffic element is part of multiple control regions, we take the mean vector as the heading direction. For example, a stop sign at a $4$-way stop intersection has $3$ control regions, one each for going straight, left turn, and right turn. The start point of the control region is the location of the traffic element itself and the end point is the location after crossing the intersection, which is different for the $3$ cases. The scene map raster $m_i$ for a node $i$ is also transformed to its own frame. %Transforming both, past trajectory and raster to node frame, simplifies the input representation for the model and accelerates model’s convergence. 
% A map encoder module $f_m$ encodes the map raster for each actor as $h^m_i = f_m\left(m_i\right)$. A fuse layer $f_l$ then combines the node type, state, and map encodings together as 
\begin{align}
    h_i = f_l\left(v_i, h_i^s, h_i^m\right)    \label{eq:encoder}
\end{align}
% where $[...]$ is the concatenation operation. 

\subsection{Discrete Interaction Prediction} \label{sec:interaction_prediction}
\vspace{-.2cm}

Supervised learning of interaction prediction requires ground-truth labels. Unlike future trajectory, discrete interaction type labels are not observed. We designed an autolabeler that computes label for actor-actor edge $(i,j)$ given their future trajectories. The autolabeler outputs: a) ${l}_{ij}$ = \texttt{IGNORING} if trajectories do not intersect, b) ${l}_{ij}$ = \texttt{GOING} if trajectories intersect and $i$ arrives at the intersection point before $j$, and c) ${l}_{ij}$ = \texttt{YIELDING} if trajectories intersect and $i$ arrives at the intersection point after $j$. We follow a semi-supervised approach where we only compute labels for actor-actor edges. Note that since traffic elements are stationary, this logic of ``who reached first'' cannot be trivially extended to element-actor edges. 
% TODO: add one line that we expect the model or our model can predict element-actor interaction based on actor-actor interaction.

Since we transformed both the past trajectory and map raster for each node in its own frame of reference, the information about the relative states of nodes is lost. This is undesirable as the model needs to be aware of the relative configuration of nodes in order to fully understand the scene. We restore this spatial information between nodes by introducing edge features for every edge that comprises the relative position and velocity of source node with respect to destination node. 

The interaction-prediction module predicts a categorical distribution over the $3$ discrete interaction types for all the edges in the graph. It is a GNN with a node model $f_{\nu}^{k}$ and an edge model $f_{\epsilon}^{k}$ for each layer $k=1,\dots,K$. The edge model first updates the edge embeddings 
\begin{align}
    e_{ij}^k = f_{\epsilon}^{k}\left(e_{ij}^{k-1}, h_i^{k-1}, h_j^{k-1}\right)
\end{align}
where $h_i^{k-1}$ is the $i^{th}$ node's embeddings. Each node then aggregates all the incoming messages (edge embeddings) and the node model updates the node embeddings as
\begin{align}
    h_i^{k} = f_{\nu}^{k}\left(h_i^{k-1}, \textstyle \sum_{j \in \mathcal{N}_i} e_{ji}^k\right)    \label{eq:interaction_prediction}
\end{align}
where $\mathcal{N}_i$ is the set of neighbors of node $i$. The initial node embedding $h_i^{0}$ is the encoder module's output $h_i$ (see Eq. \ref{eq:encoder}) Moreover, the initial edge embedding $e_{ij}^0$ is the relative state of $i$ in the $j^{th}$ node frame. An output layer $f_{ip}$ finally computes the categorical distribution as 
\begin{align}
    \hat{l}_{ij} = f_{ip}\left(e_{ij}^K\right) \label{eq:edge_output}    
\end{align}
We used a weighted cross-entropy loss over the predicted interaction type distribution. The weights are set in the inverse proportion of label distribution in the dataset. In any driving dataset, it is expected that the number of ignoring interactions will be much higher than the other two. Setting appropriate weights is necessary for preventing mode collapse.

\subsection{Traffic Light State Prediction} \label{sec:tl_prediction}
\vspace{-.2cm}

Knowledge of traffic light state is crucial for any autonomous driving system. %in order to develop legal motion plans for SDV. 
Even though traffic light locations are known in advance, their state can be either known or unknown. %due to several factors like occlusion, glare, poor visibility, etc. 
Our hybrid graph architecture enables the model to infer the state of unknown traffic light nodes by virtue of aggregating information from other nodes like state of other known traffic lights, motion of vehicles, etc. A traffic light prediction module $f_{tl}$ predicts a categorical distribution over the $3$ traffic light types (red, green, yellow) for all traffic light nodes in the graph. It takes as input the node embeddings $h_i^K$ (see Eq.~\ref{eq:interaction_prediction}) and outputs a categorical distribution $\hat{\gamma}_i = f_{tl}\left(h_i^K\right)$.

% TODO: clearly write that we are concerned only with the state of unknown traffic light and model by using 
We used a weighted cross-entropy loss over the predicted categorical distribution. Here also, we set the weights in inverse proportion to their respective class distribution. It is expected that yellow traffic light state will be less frequent than their red and green counterparts. We adopt a semi-supervised approach where the loss is computed only on known traffic light nodes during training. During inference, the model predicts the state of unknown traffic light nodes by utilizing the hybrid graph structure and sharing information along the edges.

% Note that the state of traffic light is provided as input to our model in the form of one-hot representation. So, the traffic light state prediction loss serves as an auxiliary loss for the model and forces the model to make use of the traffic light information to make accurate predictions for the vehicle’s future motion.

\subsection{Continuous Trajectory Prediction} \label{sec:traj_prediction}
\vspace{-.2cm}

We propose a new model, RecurrentGraphDecoder, to predict the future trajectory for each node in the graph. This model consists of an $N$-layer ``typed'' GNN $g$ to share information among nodes and a recurrent module $r$ to rollout the future states. Each layer $n$ comprises a node model $g_{\nu}^n$ and a separate edge model $\{g_{\epsilon}^{n,p}\}_{p=1}^{P}$ for each of the $P$ distinct interaction types, hence the name ``typed''. The edge embedding is computed as a weighted sum of the embeddings from each individual edge model 
\begin{align}
    e_{ij}^{n} = \textstyle \sum_{p=1}^P \hat{l}_{ij}^p ~ g_{\epsilon}^{n,p} \left(e_{ij}^{n-1}, h_i^{n-1},  h_j^{n-1} \right)
\end{align}
where the weights are the edge scores $\hat{l}_{ij}^p$ (see Eq.~\ref{eq:edge_output}) for the corresponding interaction type. The weighted sum allows the model to selectively attend to the different interaction types. Each node then aggregates the incoming messages and the node model $g^n_{\nu}$ updates the node embeddings as 
\begin{align}
    h_i^{n} = g_{\nu}^{n}\left(h_i^{n-1}, \textstyle \sum_{j \in \mathcal{N}_i} e_{ji}^{n}\right)    
\end{align}
where $\mathcal{N}_i$ is the set of neighbors of node $i$. We omit the time index from the above equations for clarity. The initial edge embedding $e_{ij}^0$ is the relative state of node $i$ in the $j^{th}$ node frame wheres the the initial node embedding $h_{i}^0$ is the final node embedding computed by the interaction prediction module in Eq.~\ref{eq:interaction_prediction}. 
A recurrent module $r$ predicts the next state for each node as 
\begin{equation}
\hat{s}_i^{(t)} \, , \, h_i^{(t)0} = r (s_i^{(t-1)}, h_i^{{(t-1)}{N}} )
\end{equation}
where $h_i^{{(t-1)}{N}}$ denotes the final layer node embedding at timestamp $t-1$. For predicting the states at the next timestamp $t+1$, we first compute $e_{ij}^{(t)0}$ from the predicted states $\hat{s}_i^{(t)}$ and $\hat{s}_j^{(t)}$. Since traffic elements remain stationary, we use the initial known state of the traffic elements instead of the predicted ones when computing their edge features. We then repeat this process for $N$ rounds of message passing through the typed GNN module $g$ followed by next state rollout using $r$. This formulation of predicting trajectory as a function of interaction type scores gives a high level of transparency and interpretability to our model which is crucial for developing reliable and trustworthy self-driving technologies.

% Old \tau
% where $\mathcal{N}_i$ is the set of neighbors of node $i$. The initial edge embedding $e_{ij}^0$ is the relative state of node $i$ in the $j^{th}$ node frame wheres the the initial node embedding $h_{i}^0$ is the final node embedding computed by the interaction prediction module in Eq.~\ref{eq:interaction_prediction}. 
% A recurrent module $r$ predicts the next state for each node as $s_i^{\tau+1} = r (s_i^{\tau}, (h_i^{K})^{\tau} )$ where $(h_i^{K})^{\tau}$ denotes the final layer node embeddings at timestamp $\tau$. This completes one timestamp prediction. For predicting the state of all nodes at the next timestamp, we first compute the edge features $(e_{ij}^0)^{\tau+1}$ from the predicted states $s_i^{\tau+1}$ and $s_j^{\tau+1}$ . Since traffic elements remain stationary, we use the initial known state of the traffic elements instead of the predicted ones when computing their edge features. We then repeat this process for $N$ rounds of message passing through the typed GNN module $g$ followed by next state rollout using $r$. This formulation of predicting trajectory as a function of interaction type scores gives a high level of transparency and interpretability to our model which is crucial for developing reliable and trustworthy self-driving technologies.

%$$s_i^{(\tau+1)} = r (s_i^{(\tau)}, h_i^{{K}{(\tau)}} )$$

We use Huber loss over the predicted trajectory. In order to make the model distinguish traffic element nodes from actor nodes, we force the model to predict stationary trajectory for traffic elements. The final loss function for the entire model can be written as:
\begin{align}
    \mathcal{L} = D \big(\hat{S}^{(t+1:t+T)}, {S}^{(t+1:t+T)} \big) + CE \big(\hat{L}, {L} \big) + CE \big(\hat{\gamma}, {\gamma} \big)
\end{align}
where $D$ is the Huber loss between predicted $\hat{S}^{(t+1:t+T)}$ and ground truth ${S}^{(t+1:t+T)}$ future trajectories from timestamp $t+1$ to $t+T$, $CE$ is the cross-entropy loss between predicted $\hat{L}$ and ground-truth ${L}$ interaction types and also between predicted $\hat{\gamma}$ and ground truth ${\gamma}$ traffic light states.

\section{Experiments}

\subsection{Datasets}
\vspace{-.2cm}

We used two datasets for training and evaluation purposes: 

\textbf{Internal Dataset}: We collected an autonomous driving dataset that covers two cities in the United States. We ran in-house developed perception systems to extract objects and traffic light states from the raw sensor dataset. The distribution of ignoring, going, and yielding interaction types in our dataset is $91\%$, $4.5\%$ and $4.5\%$ respectively. Our dataset has $305k$ and $80k$ instances for training and testing respectively. 

\textbf{nuScenes Dataset}: nuScenes is a large-scale autonomous driving dataset collected from the cities of Boston and Singapore. We used the manually annotated labels available with the dataset as vehicles in all the models. The distribution of ignoring, going, and yielding interaction types in this dataset is $93.2\%$, $3.4\%$ and $3.4\%$ respectively. This dataset does not have traffic light state information, but only their location, so we set all the traffic lights to be of unknown type. The nuScenes dataset contains $273k$ and $58k$ instances for training and testing respectively.

\subsection{Baselines}
\vspace{-.2cm}

We compare our proposed model, TrafficGraphNet (TGN), against three classes of methods: (a) those that do not model interaction and use only agent-centric rasters, like RasterNet~\cite{chou2019predicting}; (b) those that model interaction implicitly, like Convolutional Social Pooling (CSP)~\cite{deo2018convolutional} and Multi-Agent Tensor Fusion (MATF) ~\cite{zhao2019multi}; and (c) those that model interaction explicitly, like SpAGNN~\cite{casas2019spatially}. For MATF, we used the Multi Agent Scene version that takes in scene context and predicts a single future trajectory for each actor. For SpAGNN, we used the prediction-only part since we assume access to detected objects in a scene. 
% We refer the readers to Appendix A for further details.
We refer the readers to Appendix \ref{sec:appendix_baselines} for further details.

% NOTE: this is already in appendix. do not uncomment.
% To ensure a fair comparison of our model with baselines, we added $6$ additional channels in the raster one for each traffic element type. We also added one channel showing the current polygon of all actors in the raster. We provide the past position and velocity along with map raster as input to all the models. 

\subsection{Metrics}
\vspace{-.2cm}

We report Average Displacement Error ({ADE}) that is mean deviation between predicted and ground truth waypoints over all the horizons and Final Displacement Error ({FDE}) that is the deviation at the final horizon of $6$s. We compute these metrics for only those actors which have at least $6$s of future track. This is done to ensure a fair comparison between all the models and to make sure that all the horizons have same number of waypoints. 
We also compute the {collision rate} (CR) which is defined as the percentage of actors' predicted trajectories colliding in space-time. We generally expect that models which accurately capture actor-actor interactions will have a lower collision rate, since collisions in the real world are relatively rare (and we do not have any in our dataset).
%A model that identifies interactions properly should achieve a lower collision rate since our dataset does not contain any colliding examples.

\subsection{Results}
\vspace{-.2cm}

The implementation and training details can be found in Appendix \ref{sec:implementation_details}. The results of comparison of our model with the baselines on our dataset are shown in Table \ref{tab:comp_ours}. We observe that methods that model multi-agent interaction explicitly, TGN and SpAGNN, perform better than others. Comparing with SpAGNN, our model which explicitly represents the traffic element as nodes and reasons about their interaction with actors achieves better performance. We show the results of comparison with baselines on the nuScenes dataset in Table \ref{tab:comp_nus}. Here also, TGN achieves lowest ADE, FDE and collision rate metrics among all the methods.

\begin{table}[!h]
    \setlength\tabcolsep{4pt}
    \begin{minipage}{.5\linewidth}
      \centering
        \begin{tabular}{lccc}
        \toprule
            % Model & ADE  & FDE & CR (\%) \\
            Model & ADE (m)  & FDE (m) & CR (\%) \\
        \midrule
            TGN (ours) & \textbf{1.673} & \textbf{3.927} & \textbf{0.318} \\
            SpAGNN & 1.810 & 4.299 & 0.376 \\
            RasterNet & 2.135 & 5.199 & 0.557\\
            MATF & 2.147 & 5.200 & 0.585\\
            CSP & 2.136 & 5.219 & 0.501\\
        \bottomrule
        \end{tabular}
        \vspace{0.1cm}
      \caption{Comparison on our internal dataset.}
      \label{tab:comp_ours}
    \end{minipage}%
    \begin{minipage}{.5\linewidth}
      \centering
        \begin{tabular}{lccc}
        \toprule
            % Model & ADE & FDE  & CR (\%) \\
            Model & ADE (m) & FDE (m)  & CR (\%) \\
        \midrule
            TGN (ours) & \textbf{1.885} & \textbf{4.593} & \textbf{0.495}\\
            SpAGNN & 1.919 & 4.754 & 0.540\\
            RasterNet & 2.224 & 5.393 & 0.611\\
            MATF & 2.188 & 5.363 & 0.657\\
            CSP & 2.161 & 5.444 & 0.643\\
        \bottomrule
        \end{tabular}
        \vspace{0.1cm}
      \caption{Comparison on nuScenes dataset.}
      \label{tab:comp_nus}
    \end{minipage} 
    \vspace{-4mm} 
\end{table}

Because the vast majority of samples in our dataset are not ``interesting'' interaction scenarios, we further evaluated our model's performance by mining a set of challenging cases from our dataset. To do this, we only selected vehicles whose future average speed differs from their past average speed by at least $5m/s$, under the hypothesis that actors which must execute a significant acceleration or deceleration are more likely to be engaged in an interaction. The results of this targeted experiment are reported in Supplementary Table~\ref{tab:comp_target}. Our method significantly outperforms all baselines on this targeted test set, with a much larger performance gap than we see on the overall test set.

\subsection{Ablation Studies}
\vspace{-.2cm}

We carried out a number of ablation experiments on our dataset to understand the impact of each individual component of our model. The results of all these experiments can be found in Table \ref{tab:ablations}. 

The first set of studies is to investigate the importance of the hybrid graph formulation. In the first experiment, we investigate the effects of predicting the traffic light state. As evident from row 1, the traffic light prediction auxiliary loss slightly boosts the model's performance in terms of trajectory metrics, though not much. Next, we study the importance of having traffic elements as nodes in the scene graph as compared to stacking them as channels in the map raster, which has been done by all the prior works, to the best of our knowledge. We see a clear benefit of our hybrid graph structure compared to the traditional rasterization methods as shown in row 2.

Next, we study the importance of velocity for the model. We observe that velocity is a strong contributor in achieving low trajectory errors. Removing it from the input and output of the model leads to a drop in model performance (row 3). Moreover, there is a very substantial drop in performance when relative velocity is excluded from the edge features (row 4). This makes sense given that velocity provides motion dynamics information that plays a crucial role in motion forecasting.

% Next, we study the importance of interaction prediction module. 
% Our proposed interaction prediction module, in addition to providing transparency about the model’s prediction in a human understandable format, also helps in achieving low track error. 

Next, we replace our proposed decoder, RecurrentGraphDecoder, with other alternatives from the literature, including an RNN decoder, which predicts one timestamp at a time, and an MLP decoder, which predicts in a single-shot manner. Our RecurrentGraphDecoder, by coupling recurrent state prediction with structured information sharing between nodes at each horizon, is able to better forecast vehicles' motion compared to the other two decoders as evident from rows 6 and 8. 

Finally, we demonstrate the importance of our Interaction Prediction module in rows 5, 7 and 9. Removing this module results in a significant drop in performance, especially with the RNN and MLP decoders, which signifies the importance of information sharing between actors for accurate motion forecasting. Furthermore, even with our RecurrentGraphDecoder, we see that removing the discrete interaction prediction module harms our performance, which indicates that adding extra structure to the model by capturing semantically meaningful interaction types is also useful.

\begin{table}[]
\centering
% \setlength\tabcolsep{5.7pt} % default value: 6pt
% reduced to 4.5pt to prevent overflow due to the addition of (m) in metrics.
\setlength\tabcolsep{3.8pt} % default value: 6pt
\begin{tabular}{ccccccccc}
\toprule
& Interaction & Traffic Element & Traffic Light & \multicolumn{2}{c}{Velocity} & Decoder & \multicolumn{2}{c}{Metrics} \\
~\#~ & Prediction  & Nodes in Graph & Prediction & In/Out  & Edge  & Type & ADE (m)  & FDE (m)  \\
% \cmidrule(lr){1-1} \cmidrule(lr){2-3} \cmidrule(lr){4-4} \cmidrule(lr){5-6} \cmidrule(lr){7-9} \cmidrule(lr){10-11}
\cmidrule(lr){1-1} \cmidrule(lr){2-2} \cmidrule(lr){3-3} \cmidrule(lr){4-4} \cmidrule(lr){5-6} \cmidrule(lr){7-7} \cmidrule(lr){8-9}
0  & \cmark  & \cmark  & \cmark     & \cmark     & \cmark    & RecGraph   & {1.67} & {3.93} \\
1  & \cmark  & \cmark &   & \cmark     & \cmark    & RecGraph   & 1.68 & 3.95 \\
2  & \cmark  & \textbf{{\color{red}$*$}} &   & \cmark     & \cmark    & RecGraph   & 1.73 & 4.05 \\[1pt]
\hdashline\\[-7pt]
3  & \cmark  & \cmark & \cmark     &    & \cmark    & RecGraph   & 1.79 & 4.20 \\
4  & \cmark  & \cmark & \cmark     &    &    & RecGraph    & 2.58 & 5.45 \\[1pt]
\hdashline\\[-7pt]
5  &  & \cmark & \cmark     & \cmark     & \cmark    & RecGraph   & 1.70 & 4.00 \\
6  & \cmark  & \cmark & \cmark     & \cmark     & \cmark    & RNN   & 1.70 & 4.01 \\
7  &  & \cmark & \cmark     & \cmark     & \cmark    & RNN   & 2.15 & 5.30 \\
8  & \cmark  & \cmark  & \cmark     & \cmark     & \cmark    &  MLP    & 1.70 & 4.02 \\
9  &  & \cmark & \cmark     & \cmark     & \cmark    &  MLP  & 2.18 & 5.39 \\ 
% \thickhline
\bottomrule
\end{tabular}
\vspace{0.15cm}
\caption{An ablation study of TrafficGraphNet, conducted on our internal dataset. The ``{Interaction Prediction}'' column indicates whether we explicitly predict discrete interaction types. The ``{Traffic Element Nodes in Graph}'' column indicates whether traffic lights and signs are included as nodes in the graph. The ``Traffic Light Prediction'' column indicates whether traffic element self-supervision is applied. The ``Velocity'' columns indicate whether and how velocity information is used in the model. The ``Decoder Type'' column indicates how the trajectories are rolled out.\\ %}
%\textbf{*} \hypertarget{traffic-raster}{
%}
\textbf{{\color{red}$*$}} For the experiment where we remove traffic element nodes from our graph (row 2), we also add them to the rasters so that the model still has access to this information, albeit in a different form.
}
\vspace{-4mm} 
\label{tab:ablations}
\end{table}

\subsection{Qualitative Perturbation Experiments}
\vspace{-.2cm}

Qualitative evaluation tends to provide a deeper view into the models than simple metrics. In this experiment, we compare our model with SpAGNN, which is the best performing baseline model. We evaluate the model performance in two traditionally difficult scenes. 

\begin{figure}
    \centering
    \includegraphics[width=0.245\textwidth,]{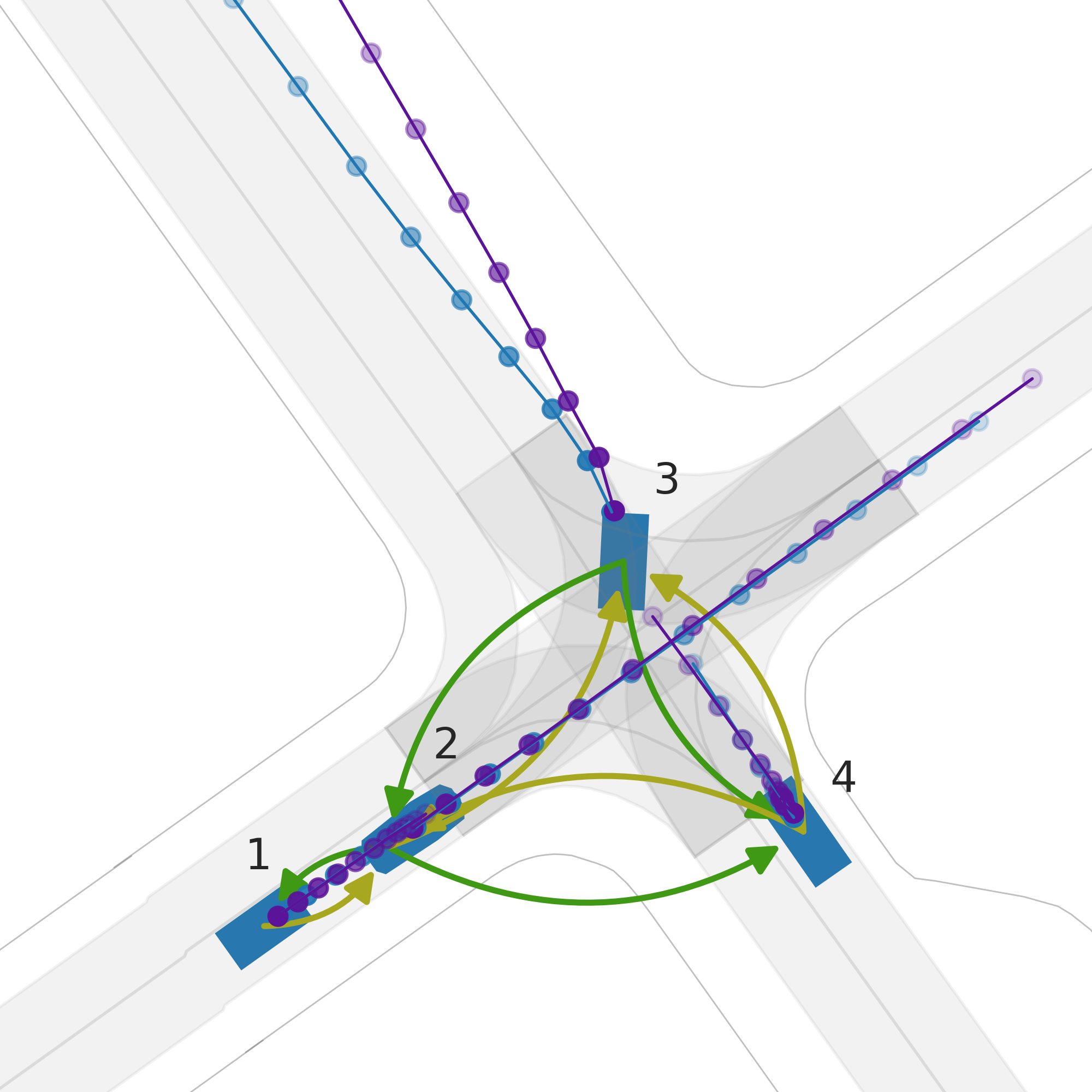}
    \includegraphics[width=0.245\linewidth,]{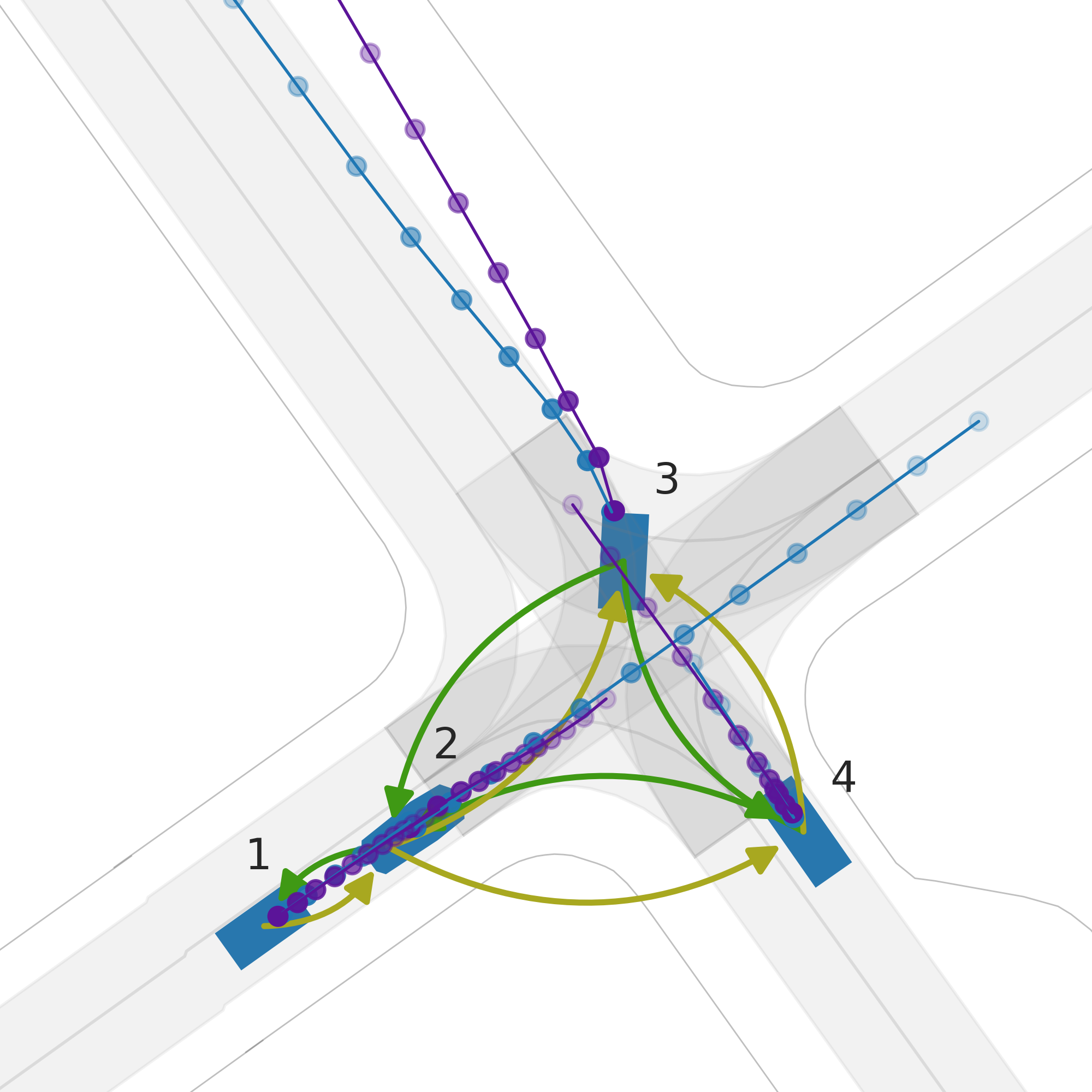} 
    \includegraphics[width=0.245\linewidth,]{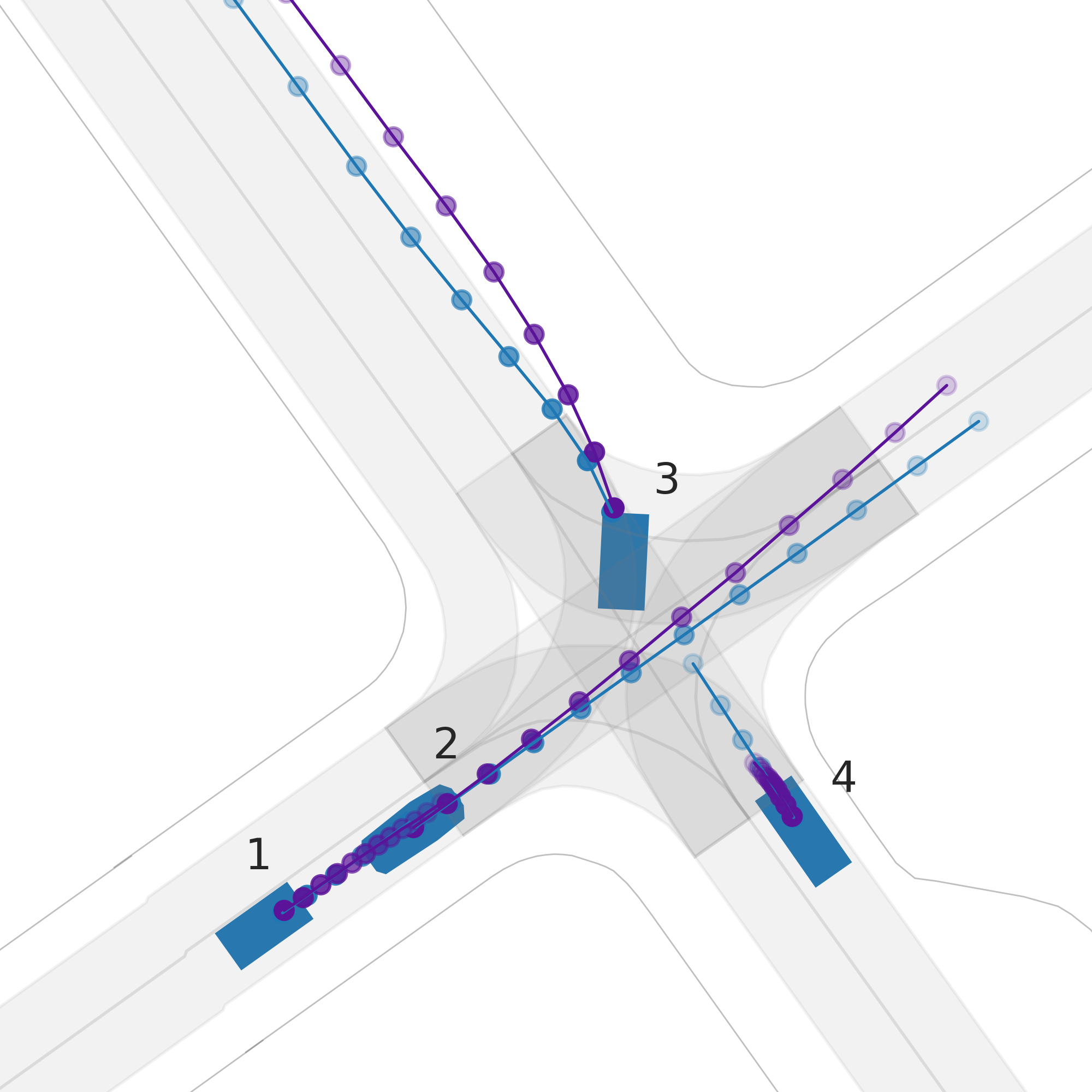}
    \includegraphics[width=0.245\linewidth,trim={2.7cm 2.2cm 2.0cm 2.7cm},clip]{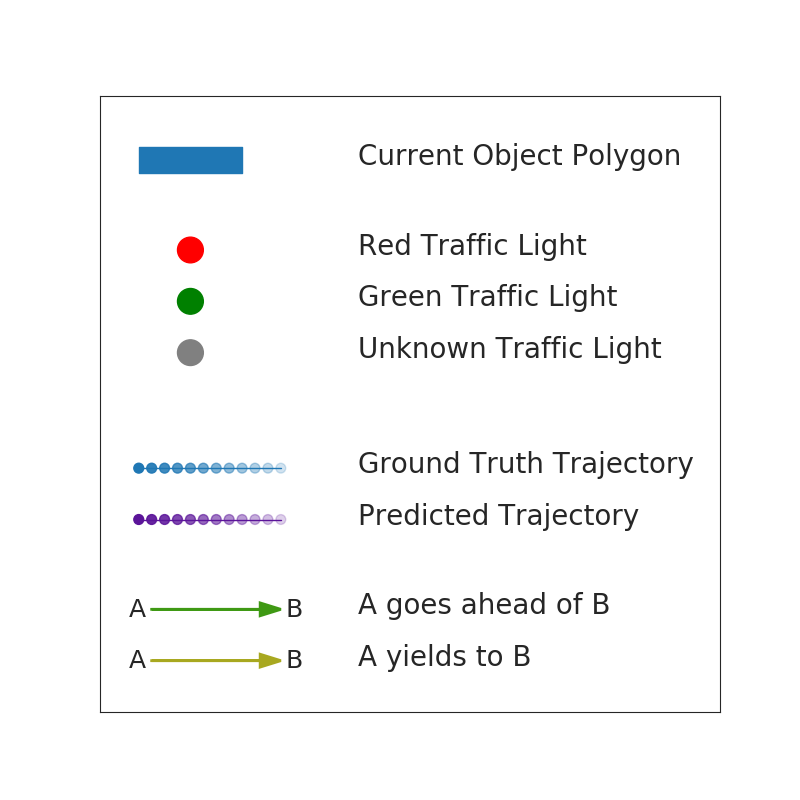} 
    \\
    \subcaptionbox{Ours, Original\label{fig:qualitative:a}}{\includegraphics[width=0.245\textwidth,]{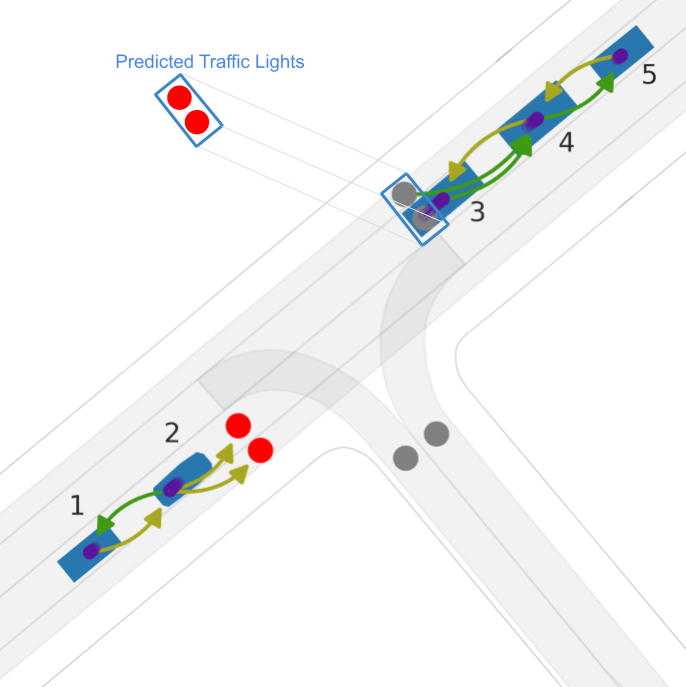}}
    \subcaptionbox{Ours, Perturbed \label{fig:qualitative:b}}{\includegraphics[width=0.245\textwidth,]{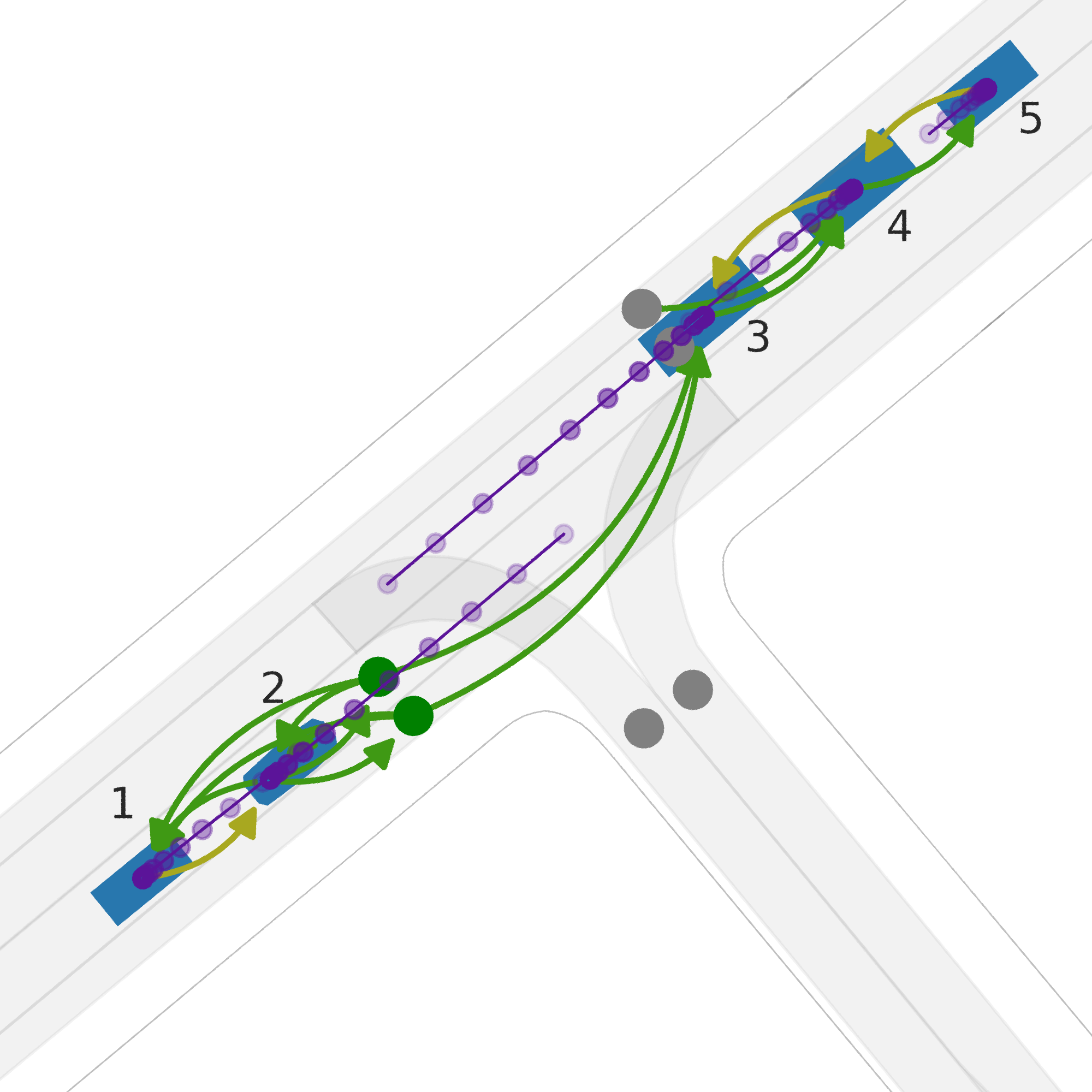}}
    \subcaptionbox{SpAGNN, Original\label{fig:qualitative:c}}{\includegraphics[width=0.245\textwidth,]{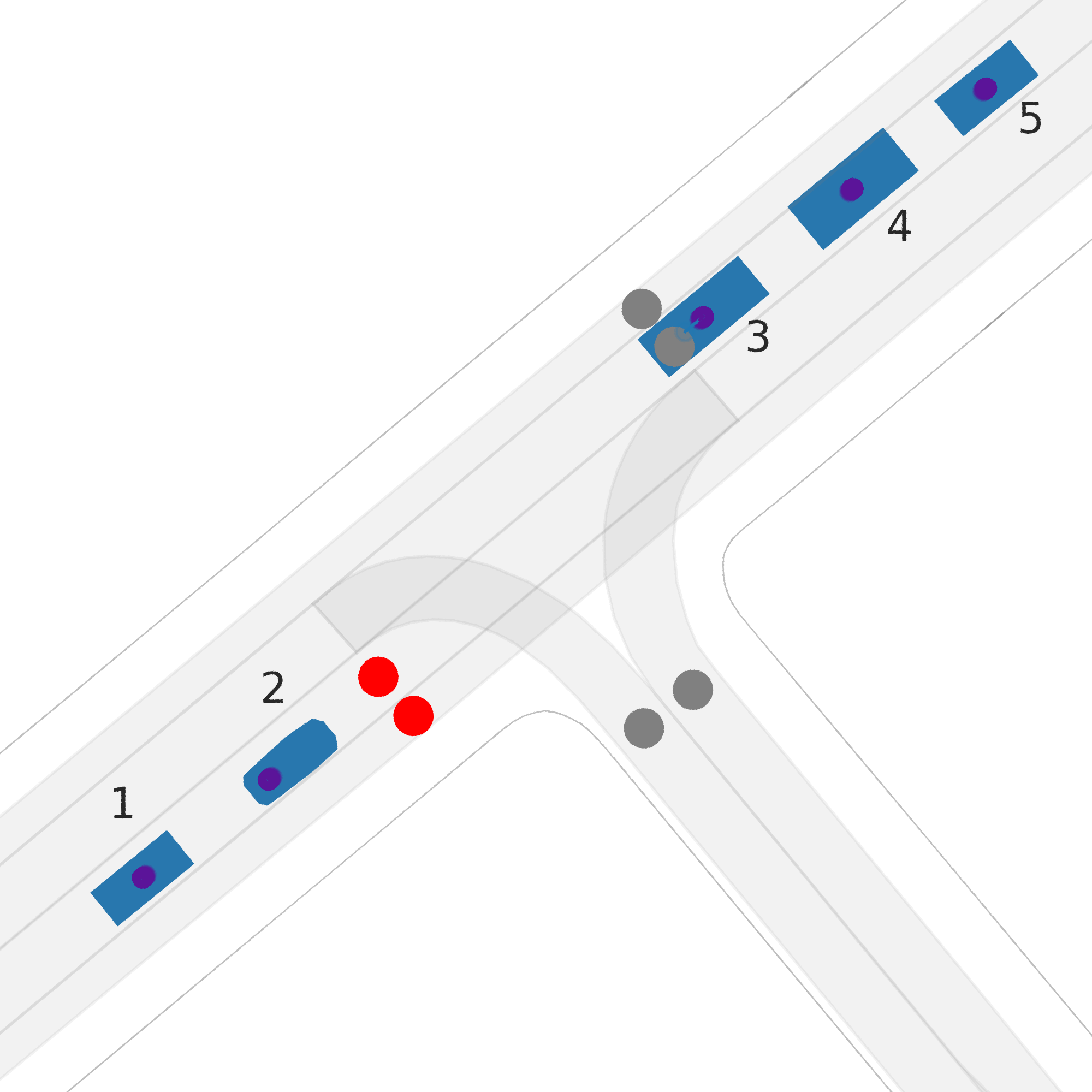}}
    \subcaptionbox{SpAGNN, Perturbed\label{fig:qualitative:d}}{\includegraphics[width=0.245\textwidth,]{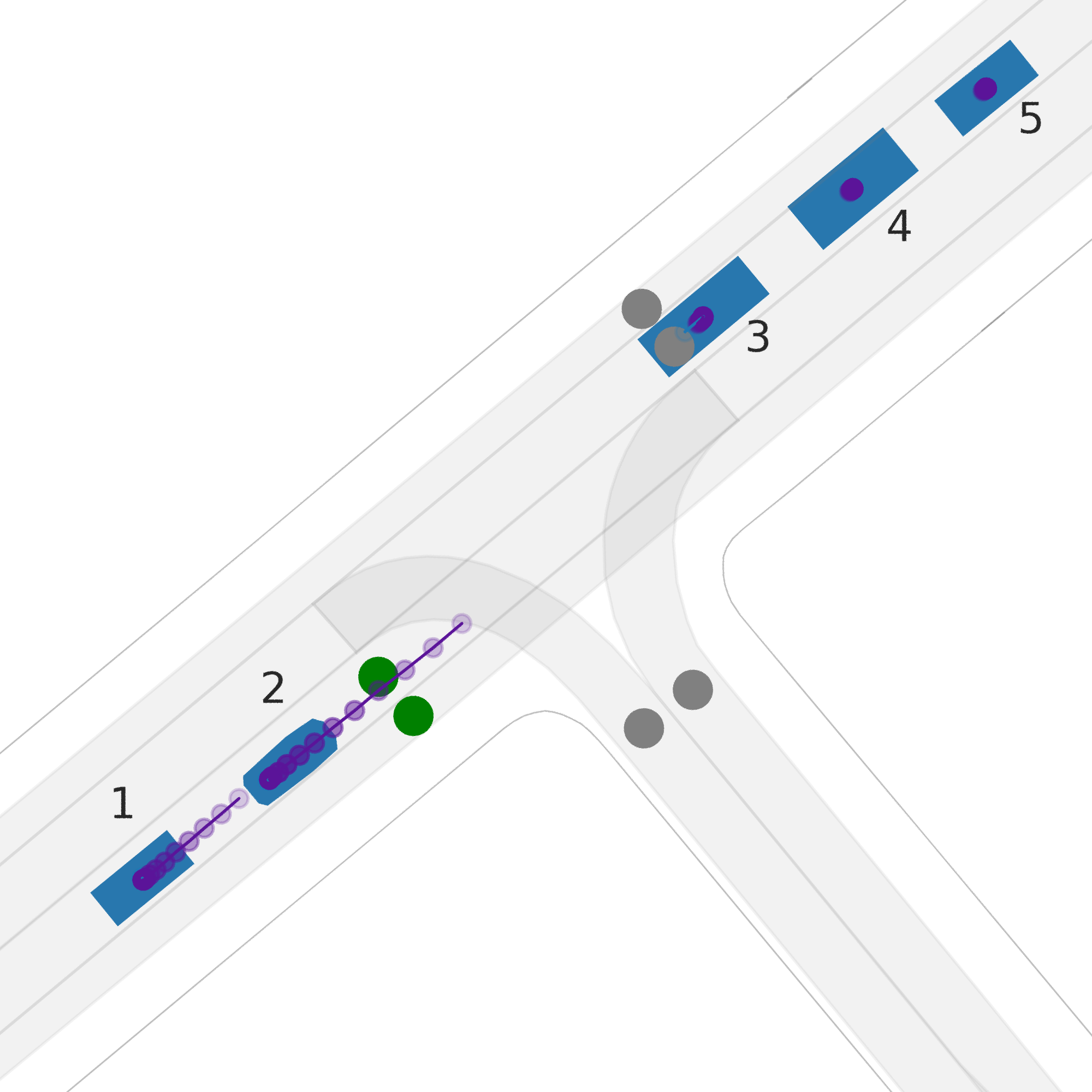}}
    \caption{Qualitative results on two interaction scenarios: a 4-way stop intersection with 4 vehicles (top), and a traffic light controlled intersection with 5 vehicles (bottom). In columns (a) and (c), we show the original predictions from our model and SpAGNN, respectively. In columns (b) and (d), we investigate how the predictions change when we manually perturb the scene. In the top row, we override the original discrete interaction prediction between actors 2 and 4, and we see their predicted trajectories change in response (after the perturbation, 2's trajectory yields to 4's). In the bottom row, we flip a red traffic light to green, and we see the predicted trajectories start to move. Note that we can only do the discrete interaction perturbation for our model since SpAGNN does not have explicit interaction types, which is why we have no result for the top row in column (d).}
    \label{fig:qualitative}
\vspace{-4mm}    
\end{figure}

The top row in Figure~\ref{fig:qualitative} illustrates a scene in a four-way-stop intersection. It can be seen from columns (a) and (c) that both our model and SpAGNN are able to generate accurate prediction that actor 4 is yielding to actor 2. However, our model offers not only continuous trajectory predictions, but also the arrows between actor 2 and actor 4 clearly indicates that actor 4 is yielding to actor 2, and actor 2 is going ahead of actor 4. To test the assumption that our model indeed learned meaningful semantics of yielding/going, we artificially perturb the scene. As can be seen in column (b) of the top row of Figure~\ref{fig:qualitative}, we artificially force the interaction prediction to change from ``actor 4 yields to actor 2'' to ``actor 2 yields to actor 4''. The resulting trajectories clearly show that the model learns to leverage those interaction predictions to develop trajectories. Moreover, we can use this technique to generate various predictions for different ``scenarios'' at the scene level, and each ``scenario'' can be assigned a probability mass that corresponds to the likelihood of the interactions.

Similarly, the bottom row of Figure~\ref{fig:qualitative} shows the predictions with traffic signals. Both our model and SpAGNN are able to predict actors will remain stopped for red lights. We again artificially perturb the scene by toggling the traffic lights on the left from red to green. As seen from columns (b) and (d), %then toggle the traffic lights on the left side from red to green (column b and d in the bottom row). 
both models predict the actors will accelerate for the green lights. However, SpAGNN is unable to predict going trajectories for the vehicles in the opposing lane that have unknown traffic lights. In contrast, our model can predict not only that actor 2 will move, but also that actor 3 will move. 

% One explanation for this is that the unknown-traffic-light prediction module in our model actually predicted the two traffic lights in front of actor 3 are green, and therefore can leverage this knowledge to predict actor 3 will go for the unknown traffic light as well.

\section{Conclusion}

In this paper we present TrafficGraphNet, a model for jointly predicting discrete interactions and continuous trajectories for a set of actors in a scene. Our model has three novel components. First, we recognize that temporal interactions with {traffic elements} such as red lights, green lights, and stop signs are just as important as interactions with other actors, and we directly include these elements as nodes in our ``hybrid'' traffic graph. Second, we not only employ a message passing approach over nodes in the graph as an information sharing mechanism, but we also introduce an explicit edge-level discrete interaction prediction which both aids our ability to predict actors' future trajectories and provides a semantic {explanation} for the predicted motion. Third, we introduce a novel RecurrentGraphDecoder architecture which jointly rolls out future actor states in a recurrent manner while sharing information over the graph at every time step. Together, these key components of our model allow us to achieve state-of-the-art trajectory prediction accuracy on both an internal dataset and the public nuScenes dataset. Beyond that, we highlight a set of perturbation experiments in which we manually intervene in the graph and are able to elicit a completely different set of trajectory predictions, indicating that our model learns a highly structured latent representation of the scene. This structure provides interpretability, which is important for real-world robotic systems. % a highly structured function that is similar 

\bibliography{references}

\appendix
\BeforeBeginEnvironment{appendices}{\clearpage}

\begin{appendices}

\section{Implementation Details} \label{sec:implementation_details}
In this section, we give a detailed description of the implementation of our model, TrafficGraphNet:

\textbf{Filtering}: We select non-parked vehicles including SDV as actors in the graph. We remove any actor further than $300$m from SDV. As mentioned before, a traffic element is connected to an actor if the distance between them is less than $25$m. If any traffic element is not connected to any actor, then it is not included in the hybrid graph.

\textbf{Input representation}: 
All models, ours and baselines, take as input past positions and velocities for the past $2$s and predict future trajectory for each actor for the next $6$s horizon. Both the input and output data have a frequency of $2$Hz. We also concatenate a $2$ dimensional boolean mask with value $1$ if that state was observed, $0$ otherwise. This is done to provide information to the model to distinguish between past state values that were padded and that were observed in reality. The current state for each actor is appended at the end of its past states. Essentially, the past states input is $(B, 5, 6)$ dimensional where $B$ is the total number of nodes in a batch, $5$ is the number of previous and current steps, and $6$ is concatenation of two-dimensional vectors of position, speed, and mask.

Our map raster consists of $7$ channels, those are: all lane boundary, left turn region, right turn region, lane all motion path, left lane marker boundary, right lane marker boundary and lane vehicle motion path. We use a spatial resolution of 0.2m/pixel and consider field of view of $30$m in front, $10$m in back, $20$m on each side for any actor. The total size of the rasters in a scene is, $(B, 7, 200, 200)$ in pixel. 

\textbf{Model architecture}:
All the models use hidden size of $128$ with ReLU activation, unless mentioned otherwise.

\textit{Hybrid Graph Representation}: The state encoder $f_s$ is a single layer RNN encoder with a GRU cell that transforms the past states input to $(B, 128)$ dimensional encoding. The map encoder $f_m$ is a simple $5$-layer CNN followed a linear layer. Each CNN layer has $32$ output channels, kernel size $3$ and stride $2$ with batchnorm. The output of CNN is $(B, 4, 4)$ feature map tensor which is flattened and fed to the linear layer which transforms into a $(B, 128)$ tensor. The fuse layer $f_l$ is a $2$-layer MLP. It takes as input the concatenated node-type one-hot, node past state encoding $h_i^s$ and node map raster encoding $h_i^m$ tensor and outputs $(B, 128)$ dimensional embedding. 

\textit{Discrete Interaction Prediction}:
Edge feature is a $4$d tensor consisting of relative position and velocity of the source node with respect to the destination node. We use $K=2$ GNN blocks in the interaction prediction module. Each block consists of one edge model and one node model. All the edge and node models are $2$ layer MLP with output size $128$. The output module $f_{ip}$ is a $2$ layer MLP. The discrete interaction prediction cross entropy loss weights are $[0.042, 1.0, 1.0]$ for our dataset and $[0.0363, 1.0, 1.0]$ for nuScenes dataset for \texttt{IGNORING}, \texttt{GOING} and \texttt{YIELDING} classes respectively.

\textit{Traffic Light State Prediction}:
The traffic light state prediction module $f_{tl}$ is a $2$ layer MLP. The cross entropy loss weights are $[0.058, 1, 0.068]$ for our dataset for red, yellow and green classes respectively. There is no traffic light prediction in NuScense dataset because traffic light states are not available.

\textit{Continuous Trajectory Prediction}:
The RecurrentGraphDecoder consists of $N=2$ TypedGNN blocks. As in interaction prediction GNNBlock, here also, each node and edge model is a $2$ layer MLP. The recurrent module $r$ is a single layer GRU. The output of $r$ is a $(B, 12, 4)$ dimensional tensor with position and velocity of each node in its own frame of reference.

\textbf{Training}:
All the models are implemented in PyTorch and trained in a distributed manner using Horovod using 4 GPUs. We use an initial learning rate of $0.001$ with Adam optimizer. All the models are trained for $20$ epochs on both datasets. 

\section{Baselines} \label{sec:appendix_baselines}
\textbf{Filtering}: We selected non-parked vehicles including SDV as actors and remove any actor farther than $300$m from SDV. This is same as the filtering in our model. 

For baselines, we add traffic element information as additional channels in order to do a fair comparison with our model, which represents traffic elements as nodes in the graph. We also add the current polygon of all the actors as one additional channel. Similar to the configuration used for our model, we use a resolution of $0.2$m/pixel for raster for the baseline models. Hence, the size of agent-centric raster for SpAGNN and RasterNet is $(14, 200, 200)$ pixels. MATF uses a single scene context. We consider the area around SDV extending $75$m in each of the $4$ directions to construct the map raster. Hence, the size of raster for MATF is $(14, 750, 750)$ pixels. CSP does not use any raster input.

All baseline models have the same past states representation as that of our model. They use a recurrent encoder $f_s$, which is same as the one in our model, to encode the past states sequence. We use a recurrent decoder for all the baseline models except SpAGNN which uses a message passing neural network to rollout future states for all the nodes in graph. 

% \textbf{Rasternet}: We used the same CNN backbone as used in the original paper to transform the raster into a $256$ dimensional raster encoding. The state encoding is concatenated with map raster encoding and fed to a recurrent decoder which outputs the future positions for all the actors. The recurrent decoder is a single layer GRU. The output is a $(B, 12, 2)$ tensor. 

% \textbf{SpAGNN}: We adopted the prediction-only part of this model as we assume access to detected objects. The state encoding is concatenated with raster encoding and rest of the details are the same as in paper.

% \textbf{MATF}:We used the publicly available implementation. 

% \textbf{CSP}: 

\section{Supplementary Results}
We also evaluated our model against baselines on a selected target set which is a subset of our internal dataset. The objective of this experiment is to compare models' performance on a set of ``interesting'' actors. Specifically, we selected only those non-parked vehicles which have a future trajectory of at least $6$s and whose future average speed differs from their past average speed by at least $5$ m/s. Essentially, actors with significant difference in their past and future motion are selected. This target set has $14k$ actors compared to $395k$ in the full dataset. The result of this comparison can be found in Table \ref{tab:comp_target}.

\begin{table}[!h]
    \centering
        \begin{tabular}{lcc}
        \toprule
            Model & ADE (m) & FDE (m)\\
        \midrule
            TGN (ours) & \textbf{4.018} & \textbf{9.248}\\
            SpAGNN & 5.010 & 11.392\\
            RasterNet & 5.584 &	13.090\\
            MATF & 5.854 & 13.777\\
            CSP & 6.037	& 14.424\\
        \bottomrule
        \end{tabular}
        \vspace{0.1cm}
      \caption{Comparison on our target dataset.}
      \label{tab:comp_target}
\end{table}

We observe that our model achieves better performance than all the baselines on this selected target set. This shows the importance of explicitly modelling interaction between agents and traffic elements in forecasting motion of ``interesting'' actors. 

\end{appendices}

\end{document}